\documentclass[letterpaper, 10 pt, conference]{ieeeconf} 
\IEEEoverridecommandlockouts                              

\usepackage{amsmath}
\usepackage{amssymb}
\usepackage{balance}

\usepackage{cite}
\usepackage{algorithmic}
\usepackage{graphicx}
\usepackage{textcomp}
\usepackage[hidelinks]{hyperref}
\usepackage[table]{xcolor}

\title{\LARGE \bf
Dynamic Human-Robot Role Allocation based on \\ Human Ergonomics Risk Prediction and Robot Actions Adaptation
\thanks{This work was supported by the European Research Council's (ERC) starting grant Ergo-Lean (GA 850932).}
}

\author{Elena Merlo$^{1,2}$, Edoardo Lamon$^{1}$, Fabio Fusaro$^{1,3}$, Marta Lorenzini$^{1}$, Alessandro Carfì$^{2}$,\\ Fulvio Mastrogiovanni$^{2}$, and Arash Ajoudani$^{1}$
\thanks{$^1$ Human-Robot Interfaces and physical Interaction, Istituto Italiano di Tecnologia, Genoa, Italy. \tt\small edoardo.lamon@iit.it}
\thanks{$^2$ Dept. of Informatics, Bioengineering,
Robotics, and Systems Engineering, University of Genoa, Genoa, Italy.}
\thanks{$^3$ Dept. of Electronics, Information and Bioengineering, Politecnico di Milano, Italy.}
}

\begin{document}

\maketitle
\thispagestyle{empty}
\pagestyle{empty}

\begin{abstract}
Even though cobots have high potential in bringing several benefits in the manufacturing and logistic processes, their rapid (re-)deployment in changing environments is still limited. To enable fast adaptation to new product demands and to boost the fitness of the human workers to the allocated tasks, we propose a novel method that optimizes assembly strategies and distributes the effort among the workers in human-robot cooperative tasks. The cooperation model exploits AND/OR Graphs that we adapted to solve also the role allocation problem. The allocation algorithm considers quantitative measurements that are computed online to describe human operators' ergonomic status and task properties. We conducted preliminary experiments to demonstrate that the proposed approach succeeds in controlling the task allocation process to ensure safe and ergonomic conditions for the human worker. 
\end{abstract}

\section{INTRODUCTION}
Human-Robot Collaboration (HRC) envisions humans and robots not only to coexist in a fenceless environment but also, while working simultaneously in the same workspace, to share tasks and goals. Robotic platforms capable to collaborate with humans (cobots) are convenient especially in scenarios of small and medium-sized enterprises (SMEs), characterized by flexible and agile manufacturing requirements, since cobots, compared to the highly specialized platforms, are fast re-configurable and adaptable to the new product demand. Moreover, the close collaboration unlocked by cobots allows pairing robot qualities (e.g., endurance and precision) with human ones (e.g., flexibility and experience) to improve production efficiency. Within this scenario, it is possible to allocate heavy and repetitive tasks to cobots to avoid work-related musculoskeletal disorders (WMSDs), which still represent the biggest problem in terms of absenteeism, and, hence, lost productivity among workers in industries~\cite{govaerts2021prevalence}.
The integration of ergonomics principles in the design of collaboration strategies has the potential to free human operators from risky tasks, which affect the entire process in terms of time and costs~\cite{cherubini2016collaborative, villani2018survey, ajoudani2018progress}.

In this manuscript, we will investigate the development of a new method to ensure an ergonomic and fruitful HRC, that profits from quantitative measurements, such as the human kinematic state and task characteristics.
Such measures (e.g. human posture and muscle activation) can be integrated both in the task planning process~\cite{johannsmeier2016hierarchical, busch2018planning, el2019task, lamon2019capability} and in the control loop~\cite{kim2017anticipatory, busch2017postural, peternel2018robot}. The recent advances of technology in the hardware design of torque-controlled robots have enabled the development of control algorithms, such as impedance and admittance control \cite{albu2007unified}, that allow cobots to perform a vast number of tasks. 
However, to optimize the teamed performance, collaborative control strategies tend to overfit the specific task, and it is not simple to re-use the same algorithm for different tasks. For instance, in the collaborative solution implemented by \cite{peternel2018robot}, the physical behavior of the robot is adapted online to the human motor fatigue, measured through EMG sensors, placed at the human shoulder joint. However, according to the executed task (material sawing or surface polishing), the control policy for optimizing the performance changes.
On the other hand, embedding the teamed performance optimization at the planning level allows to apply the same architecture for many different tasks and to regulate the worker's effort distribution monitoring the human physical status at a lower frequency.
For these reasons, we will focus on including ergonomics principles in the design of task planning and role allocation algorithms.

\begin{figure*} [!t]
    \centering
	\includegraphics[trim=0.0cm 0.0cm 0.0cm 0.0cm,clip,width=1\linewidth]{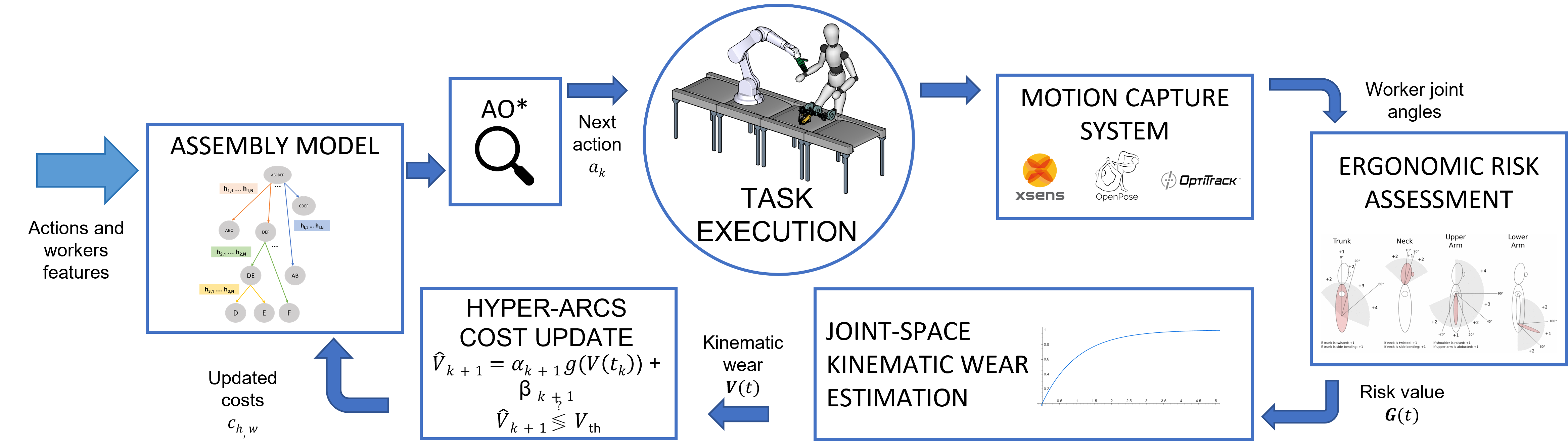}
    \caption{Framework scheme. At each iteration, the algorithm returns the next task and the allocated worker. After the execution of the task by the human, the joints-level ergonomic risk and, hence, the \textit{kinematic wear} value $V_i(t)$, that contains the kinematic history of each joint, are evaluated. According to that, the AOG hyper-arc costs are updated for the next allocation.}
	\label{fig:aog_scheme}
	\vspace{-0.3cm}
\end{figure*}
Johannsmeier \& Haddadin, for instance, solve the role allocation problem through an A* search on a AND/OR Graph (AOG) exploiting the arc-related costs, modified to account for the agents' dissimilarities~\cite{johannsmeier2016hierarchical}.  
The indexes that should be measured to thoroughly capture the task-worker suitability are presented by Lamon et al.~\cite{lamon2019capability}.
Such metrics describe the agents' nature, in terms of kinematic and dynamic characteristics, and their skills related to the task properties. In particular, agent effort considers both ergonomics and safety constraints.
The main limitation of these approaches is that the optimization runs offline and, hence, the allocated tasks do not change over time according to the human physical status. For instance, some heavy tasks might be feasible for the human workers when they are rested, but, after a few repetitions, the performance of the same task might impose a large risk to workers' health. 
To account for the task variability, the work presented by Darvish et al, 2021 pre-computes offline all the possible assembly sequences. Then, online, the human worker can execute the suggested action or delegate it to the cobot. In the latter case, the AOG, acting as task planner, automatically switches to a new branch~\cite{darvish2021hierarchical}. 
However, the actions are all allocated beforehand; then the allocation results remain fixed for the whole teamwork. Moreover, for complex assemblies, keeping all the possible assembly sequences might not be feasible, due to limited storage capacity.
Other methods have been developed to dynamically schedule and allocate tasks to workers by minimizing other indexes, such as the overall execution time \cite{casalino2019optimal, fusaro2021integrated}. Nevertheless, these methods cannot account for the variety of different assembly sequences that complex tasks might impose.

To overcome the limitation of the state-of-the-art methods, we propose an online role allocation strategy, that can assign actions among the agents of the team according to the physical human-worker status. The main contribution of the method is twofold:
\begin{itemize}
    \item The introduction of online role allocation within the AOG framework, which, at each step of the assembly task, provides the next action and the allocated worker, as a result of an optimization algorithm AO*. 
    \item The integration of human joint-level status indicator, which we called \textit{kinematic wear}, can account for the usage of each joint during the execution of an assembly task of lightweight pieces. Such an indicator can be estimated with different ergonomics assessment methods. Nevertheless, the framework can be paired also with other metrics representing human risk, task performance, etc. 
\end{itemize}
The performances of the method are investigated with simulations that aim to evaluate its time complexity, and with a proof-of-concept assembly. The results show the potential of the strategy in preventing risky actions during human-robot cooperative industrial tasks.

\section{AOG FOR INDUSTRIAL TASK PLANNING}
A commonly used approach for representing well-structured industrial tasks, such as assemblies, exploits \textit{AND/OR Graphs} (AOGs). AOGs model all the possible assembly sequences of an assembly task in a compact representation with fewer nodes compared to the general Directed Graphs~\cite{de1990and}. 
Recently, researchers extended the formulation of AOGs to embed also the role allocation problem in human-robot assembly tasks. With such a method, not only the assembly sequence but also the task-agent pairing can be optimized~\cite{johannsmeier2016hierarchical}.
An AOG is a data structure characterized by a set of nodes $N=\{n_1, n_2, \dots, n_{|N|}\}$ and a set of hyper-arcs $H=\{h_1, h_2, \dots, h_{|H|}\}$. Each node $n \in N$ represents a state of the decomposed task, while hyper-arcs define the transitions between states. Each $h \in H$ describes a many-to-one transition, meaning that it connects a set of child nodes with a father node. In the case of assembly tasks, considering that most of the assembly operations join two sub-assemblies, hyper-arcs are modeled as two-to-one connectors. 
The child nodes connected by the same $h$ are in a logical AND, while different hyper-arcs with the same parent node are in logical OR. The only node without a father is named root. The nodes without children are identified as leaf nodes.

To embed the role allocation problem in the AOG formulation, two additional sets are defined: the set of workers involved in the collaboration $W=\{w_1, w_2, \dots, w_{|W|}\}$ and the set of assembly actions $A=\{a_1, a_2, \dots, a_{|A|}\}$ that workers have to perform. From now on, actions depict both proper assemblies (e.g. screwing two pieces together) and `relaxed' assemblies (e.g. moving an object on top of a table could be considered as an assembly between such an object and the table). 
The desired assembly sequence among all the possible ones the AOG describes can be computed through the assignment of a cost to each $h \in H$ ($c_{h_1}, c_{h_2}, \dots, c_{h_{|H|}}$). Such values can encode the complexity of performing each assembly action, and, by exploiting an optimal-based search algorithm, the path with the minimum total cost can be found. In a cooperative scenario, the same action can have a different cost for each worker. Therefore, to obtain the optimal sequence both in terms of task characteristics and workers' skills, each hyper-arc is repeated for $|W|$ times and a cost, that represents the suitability of $w_i$ to that action, is assigned to each of them.

\section{ERGONOMIC ROLE ALLOCATION}
To retrieve the desired assembly sequence and the optimal action allocation, a custom AO* search is implemented, since, unlike the standard AO*, the goal of the algorithm is to inspect all the leaf nodes, as they represent the atomic pieces which are all used in the assembly.
Therefore, the search acts in a top-down fashion, from the root node, which represents the state where all the pieces are assembled, to the leaf nodes.
The desired path is the one that minimizes the sum of the costs of the traveled hyper-arcs.
While the optimality of the solution, with fixed costs, is ensured, if costs change during the task execution, the search algorithm should be executed after the completion of each action. In such a case, it explores a reduced graph, from the root node to the last reached state.
This online mechanism is exploited in our framework (see \autoref{fig:aog_scheme}). For simplicity, the costs for the actions of robot $w_r$ are fixed, $c_{h_j,w_r} = c_{R}$. Instead, costs for human actions are computed according to a joint-level prediction of the ergonomic risk associated with the execution of an action by a human worker. 
Such a vector should not only acknowledge instantaneous non-ergonomic postures but also identify situations where the human worker stands for a long time in risky configurations. For this reason, the concept of \textit{kinematic wear} $V(t)$ is introduced.
The kinematic wear level should be a continuous function, i.e. the initial condition of kinematic wear of the next action $a_{k+1}$ is the accumulated kinematic wear $V_k(t)$ at the end of the previous action $a_{k}$ and limited between $V_{min}$ and $V_{max}$. In general, different models of kinematic (and also dynamic) wear can be used. The authors of the manuscript exploited in other works dynamic indicators to regulate the human-robot interaction \cite{lamon2019capability, lorenzini2019new} in high-force demanding tasks. In this work, instead, assemblies of a large number of lightweight pieces are addressed, and, hence, a kinematic index can capture more precisely potentially damaging situations.  
After the execution of each action $a_k$ by a human worker, the \textit{kinematic wear} vector $\boldsymbol{V}(t_k) = \begin{bmatrix}V_{1}(t_k) & ... & V_{m}(t_k)\end{bmatrix}^T$ is computed, where $m$ is the size of the monitored joints.

To predict the ergonomic risk associated with the execution of all the possible $a_{k+1}$ actions, we assume the following:
\begin{itemize}
    \item[(i)] the prediction model is linear to the initial conditions. i.e. the wear accumulated by the previous actions;
    \item[(ii)] the increase due to the execution of action $a_{k+1}$, i.e. the model parameters, does not depend on the past actions. 
\end{itemize}
The prediction $\hat{V}(t_{k+1})$ of the risk associated with the execution of the next action $a_{k+1}$, where $t_{k}$ is the instant when $a_{k}$ was completed, can be computed as:
\begin{equation} \label{eq:kin_wear_pred}
    \hat{V}(t_{k+1}) = \alpha_{k+1}\;g({V}(t_{k})) + \beta_{k+1}
\end{equation}
where $\alpha_{k+1}$ and $\beta_{k+1}$ are the linear parameters that regulate the raise of \textit{kinematic wear} over time and $g(V(\cdot))$ is a function of the initial conditions of $V(\cdot)$.
Moreover, $\alpha_k$ and $\beta_k$ are representative of the level of the joint involvement in the action $a_k$ and hence should be estimated offline.

With model \eqref{eq:kin_wear_pred} it is possible to predict the risk associated to the execution of each action by the human worker. Within $[V_{min},V_{max}]$, three different risk ranges are defined: \textit{low}, \textit{medium}, \textit{high}.
While low means negligible risk, high means that the joint is in a potentially damaging condition. The hyper-arc cost is designed according to the range to which the prediction of each joint \textit{kinematic wear} belongs:
\begin{equation} \label{eq:kin_wear_ranges}
    \gamma_i =
    \begin{cases}
    \gamma_{high} \qquad &\text{if} \quad \hat{V}_{i}(t_{k+1}) \geq  V_{{th}_2}, \\
    \gamma_{med} \qquad &\text{if} \quad  V_{{th}_1} < \hat{V}_{i}(t_{k+1}) <  V_{{th}_2},\\
    \gamma_{low} \qquad &\text{if} \quad \hat{V}_{i}(t_{k+1}) \leq V_{{th}_1}.
    \end{cases}
\end{equation}
where $\gamma_{high} > \gamma_{med} > \gamma_{low}$. Finally, the cost $c_{h_j,w_h}$ associated to hyper-arc $h_j$ that represents action $a_k$ executed by a human worker $w_h$ is
\begin{equation} \label{eq:human_cost}
    c_{h_j,w_h} = \sum_{i=1}^{m} \gamma_{i}.
\end{equation}
Instead of this simple sum, it is possible to introduce a weighted sum with a different weight for each joint, to ensure that some joints prevail in the cost computation.

\section{EXPERIMENT}
To assess the proposed approach we tested its performances in terms of computational time and results of allocation in a proof-of-concept cooperative industrial assembly task. Both experiments were carried out on a laptop with an Intel Core i7-8565U 1.8 GHz × 8-cores CPU and 8 GB RAM. The architecture has been developed in C++ and Python, on Ubuntu 18.04 and ROS Melodic.
\subsection{Computational Complexity Evaluation}
\begin{figure}
    \centering
    \includegraphics[trim=2.5cm 0.5cm 4.5cm 1.5cm,clip,width=\linewidth]{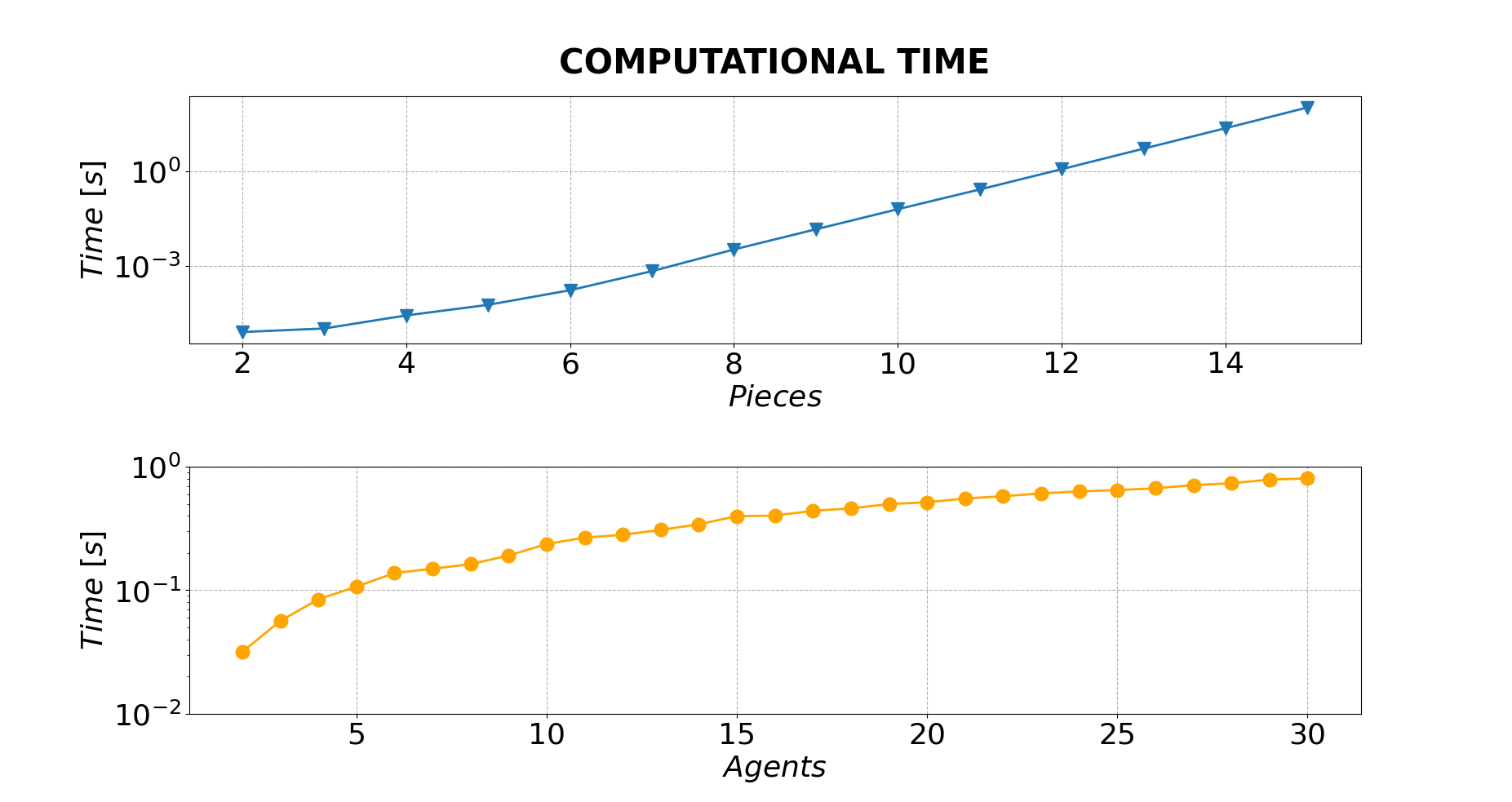}
    \caption{AO* computational time (in log scale) obtained by increasing the number of pieces to be assembled (top) and workers (bottom).}
    \label{fig:computational_time}
    \vspace{-2mm}
\end{figure}

First, since the search is iteratively performed after the execution of each action, it is important to ensure that a solution can be found within the duration of the current action. The computational complexity of the AO* algorithm was evaluated by running the search on AOG graphs with an incremental size of:
\begin{itemize}
    \item[(i)] leaf nodes, from 2 to 15, representing the atomic assembly pieces (with a fixed number of agents equal to 2).
    \item[(ii)] agents involved in the cooperation, from 2 to 30 (with a fixed number of assembly pieces equal to 10).
\end{itemize}
These AOGs were generated by considering the hypothesis of stable and feasible interconnections between adjacent pieces. This means that, given N pieces, there are N-1 interconnections between the N pieces, i.e. the i-th interconnection connects piece $p_i$ and part $p_{i+1}$~\cite{de1990and}. 
In \autoref{fig:computational_time} both graphs present an exponential trend, comparable with the ones presented in~\cite{darvish2018flexible}. In particular, the plot in the bottom graph (increasing number of agents) presents a flattened curve. This is because an increment of the number of agents implies a raise in the number of hyper-arcs, while an increment of the number of pieces entails a raise of both nodes and hyper-arcs.
An assembly with 15 pieces and 2 agents corresponds, in the worst case, to an AOG with 120 nodes and 1120 hyper-arcs, and the optimal search on such a graph is a considerably large problem. Anyway, the hypothesis that all the interconnections between pieces are feasible is a severe assumption for a general assembly, i.e. not all the adjacent pieces can be connected. For example, to assemble a table, made of a plate and 4 legs, the legs cannot be assembled but only with the plate.  
Moreover, it can be noticed that, after the execution of an action, the search algorithm operates on an AOG with a reduced size, hence the time to generate iteratively the solution is exponentially descending.  

\subsection{Experimental Setup}

\begin{figure}[t]
	\centering
    \includegraphics[trim=0.0cm 0.0cm 0.0cm 0.0cm,clip,width=0.9\linewidth]{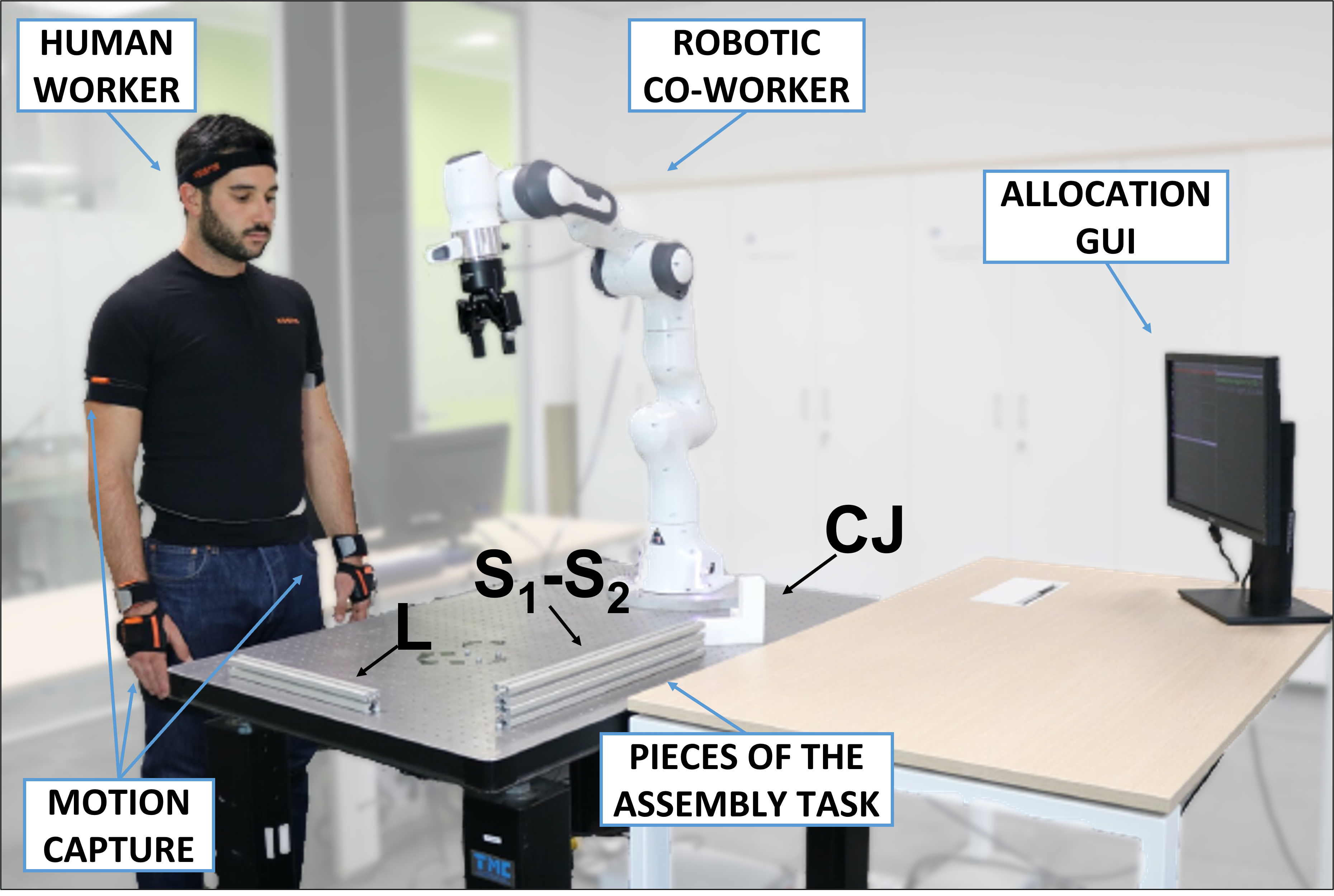}
	\caption{Experimental setup. The human worker and the robot co-worker share the workbench where the assembly pieces are placed. The worker wears the Xsens suit to capture his movements. A GUI displayed on the monitor informs the human worker of allocation results.}
	\label{fig:aog_exp_setup}
    \vspace{-2mm}
\end{figure}

To evaluate the allocation results we defined a proof-of-concept assembly task that consists of the assembly of a corner joint $CJ$ with three aluminum profiles, which are two sides $S_1, S_2$, and a leg $L$. The profiles $S_1$ and $S_2$ have the same length, while $L$ is shorter. Each of them interlocks in a predefined hollow of the corner joint.
The four pieces are placed on top of a workbench in a workspace shared by the human subject with a Franka Emika Panda manipulator, which is fixed at the operator's left.
The two aluminum profiles $S_1$ and $S_2$ are placed, one above the other, in front of the human worker, but on the opposite side of the table, as well as the corner joint $CJ$. The profile $L$, instead, is on the right side of the table. On another table, placed near the robot workspace, a monitor is located, to show to the operator the result of the allocation algorithm.
In \autoref{fig:aog_exp_setup} the experimental setup is presented. 

The atomic actions of the task are: ($a_1$) pick the corner joint $CJ$ from its initial position and place it in a pre-defined location on the workbench, ($a_2$) pick and assemble $S_1$ with $CJ$, ($a_3$) pick and assemble $S_2$ with $CJ$, ($a_4$) pick and assemble $L$ with $CJ$, ($a_5$) pick the complete assembly and place it on the additional table, further from the human. While actions $a_1$ and $a_5$ are fixed as the first and the last of the assembly sequence, the order of the other three actions is online decided by the AO*. In this experiment, the assembly sequence was fixed. The actions order was the following: $a_1$, $a_2$, $a_3$, $a_4$, and $a_5$. The variability here is introduced by the results of the dynamic allocation, computed according to the time-varying hyper-arcs costs.
A motion capture system (Xsens suit \url{xsens.com}) is in charge of capturing the human kinematic configurations, in terms of joints relative angles and links positions. The joint values are then processed by the ergonomic risk assessment method.
In this experiment, we make use of Rapid Upper Limb Assessment (RULA). It associates to each joint a positive discrete score that represents the associated postural risk: the higher the score, the higher the risk. The joints involved in the assessment method are shoulder, elbow, and wrist of the strong arm, trunk, and neck~\cite{mcatamney1993rula}.

To model the \textit{kinematic wear} index at the joint level we chose an RC circuit behavior:
\begin{equation}
    V_i(t) = 1 - (1-V_i(0))\hspace{1mm} e^{- \int_0^t{\frac{G_i(\tau)}{C}d\tau}} \quad 0 \le t \le t_{k+1}
\label{eq:kin_wear}
\end{equation}
where $V_i(t) \in [0,1)$ and $V_i(0) = V_i(t_{k})$. $C$, and $G_i(t)$ are the \textit{kinematic wear} level, the endurance capacity, and the current RULA score of the \textit{i-th} upper body joint, respectively. In particular, to ensure the continuity of the ergonomics assessment method, $G_i(t)$ is modeled as a weighted sum of sigmoid functions.
Such model has been used already in literature to describe human muscle usage~\cite{ma2010new} and also thermal motor usage~\cite{urata2008thermal}. In a similar way, with \autoref{eq:kin_wear} we would like to provide a joint-level kinematic usage descriptor.

Given that a subject, in a static configuration, can exert a low force (e.g., holding a lightweight object) for $240$ $s$ without physical discomfort~\cite{grandjean1988fitting}, the capacity $C$ is set to allow the human joint to reach, starting from $0$ initial condition, the asymptotic value of $V=1$.
To retrieve the value of $C$, \autoref{eq:kin_wear} is inverted with a value of $V_i(t) = 0.993$ (corresponding to five time constants), with an average level of RULA (risk value $G_{avg} = 3$).
The score $G_{avg} = 3$ represents a medium joint risk since the values of $G_i(t)$ are bounded by 1 to 7.
In this way, the capacity value is the same for all the joints.
\begin{equation}
    C = - G_{avg} \hspace{1mm} \frac{240}{\ln(0.007)}.
\label{eq:C_comp}
\end{equation}

In practice, according to \autoref{eq:C_comp}, the $i-th$ joint motion during the task execution generates a change of the RULA score $G_i(t)$, which, in turn, increases the corresponding kinematic wear $V_i(t)$, with a slope corresponding to the risk level of the new posture.\\
According to the cooperation protocol, we impose that while the robot executes an action, the following action cannot be allocated to the human: during this rest period the kinematic wear level of each joint decreases according to the recovery function (RC circuit discharge):
\begin{equation}
    V_i(t) = V_i(0)\hspace{1mm} e^{- \frac{r}{C}t} \quad 0 \le t \le t_{k+1}
\label{eq:recovery}
\end{equation}
where $V_i(0) = V_i(t_{k})$ and $r$ is the recovery rate. 
We set it empirically to $r=3$ which is approximately the value that allows a joint to fully recover in a period of $240$ $s$, i.e. to match the recovery time (discharge) with the wear time (charge).
By comparing model \eqref{eq:kin_wear} with \eqref{eq:kin_wear_pred}, it is straightforward to understand that the RC circuit-like model \eqref{eq:kin_wear} provides also a good prediction of the kinematic wear \eqref{eq:kin_wear_pred}, since $g({V}_i(t_{k})) = {V}_i(t_{k})$, $\alpha_{k+1,i} = e^{- \int_0^{t_{k+1}}{\frac{G_i(\tau)}{C}d\tau}}$, and $\beta_{k+1,i} = 1 - e^{- \int_0^{t_{k+1}}{\frac{G_i(\tau)}{C}d\tau}} = 1- \alpha_{k+1,i}$. As a result, only one value per joint $i$ per action $k+1$ (i.e. $\alpha_{k+1,i}$) should be estimated and stored to compute $\hat{V}_i(t_{k+1})$.

The $\gamma_i$ scores can be found in \autoref{table:aog_gamma},
\begin{table}[t]
\caption{}
\vspace{-4 mm}
\label{table:aog_gamma}
\begin{center}
\begin{tabular}{|c|c|c|}
\hline
 $\gamma_i$ & \textbf{Risk Level} & \textbf{Condition} \\
\hline
1 & LOW & $\hat{V}_{i}(t_{k+1}) \leq V_{{th}_1}$  \\
\hline
10 & MEDIUM & $V_{{th}_1} < \hat{V}_{i}(t_{k+1}) <  V_{{th}_2} $\\
\hline
100 & HIGH & $\hat{V}_{i}(t_{k+1}) \geq  V_{{th}_2},$ \\
\hline
\end{tabular}
\end{center}
\vspace{-6mm}
\end{table}
where $V_{{th}_1}=0.25$ and $V_{{th}_2}=0.75$, which represent $25\%$ and $75\%$ of the maximum value of $V_i$.
The cost of hyper-arcs modeling robot actions is experimentally fixed to $c_R = 35$, in such a way tasks can be allocated to the robot not only if the predicted $\hat{V}_i > V_{{th}_2}$ (high risk), but also in case the wear of at least 4 joints belongs at the same time to the middle range (medium risk).
These values were experimentally chosen according to the used model, in order to classify and face different risk conditions. Further studies will investigate the tuning of these parameters.

\subsection{Parameter Calibration Procedure}
\begin{figure}[t]
    \vspace{0.4cm}
	\centering
    \includegraphics[trim=0.5cm 0.0cm 4cm 0.0cm,clip,width=\linewidth]{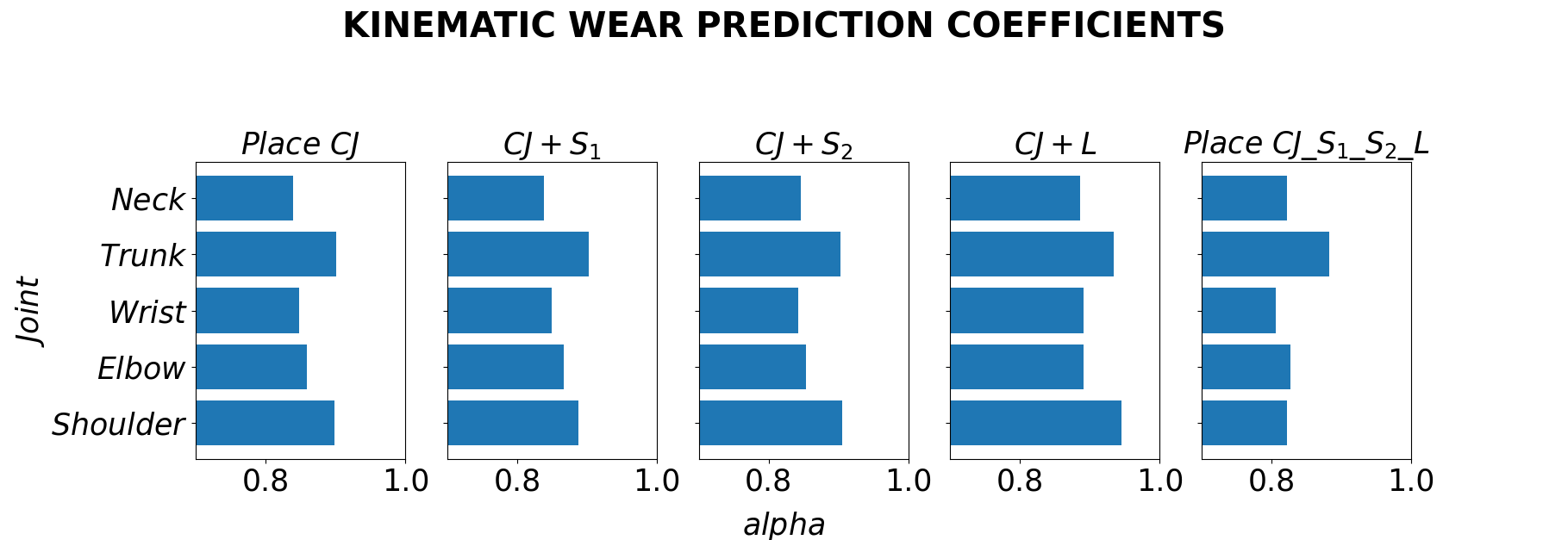}
	\caption{Joint-level \textit{kinematic wear} prediction coefficient $\alpha$ estimated in the calibration phase. The lower the $\alpha$, the higher the ergonomic risk associated.}
	\label{fig:aog_exp_alpha_calib}
    \vspace{-2mm}
\end{figure}

The experiment envisions a calibration protocol useful to find the prediction coefficient $\alpha$. 
To obtain them, the human subject is asked to perform for 10 times the same assembly action. For each repetition, we compute $\alpha$ for each joint. The desired $\alpha$ is the average between the sampled values (see \autoref{fig:aog_exp_alpha_calib}).
It is interesting to notice that, the first three actions ($a_1, a_2, a_3$) present similar coefficients: for all of them the operator is asked to pick a piece on the opposite side of the table, then the performed reach and pick movements are the same for all of them. The action $a_4$ stresses less all the joints: the higher $\alpha$ the smaller the accumulated risk in the execution of that action, due to the model we chose for \textit{kinematic wear}.
This is because $L$ lies on the worker's right side, so reach and pick movements require a lower effort. In particular, the shoulder joint is the less involved one, since it covers a small angle, remaining in the portion of its range of motion with the lowest risk. The last action $a_5$ is the most wearing since the human has to stretch to place the full assembly on the additional table. Moreover, this action demands more time, since the worker has to move both to reach the second table and to come back.

\subsection{Experimental Results}
\begin{table}[!t]
\caption{}
\vspace{-4 mm}
\label{table:allocation_res}
    \begin{center}
    \begin{tabular}{|c|c|c|c|c|c|}
    \hline
     & $\boldsymbol{a_1}$ & $\boldsymbol{a_2}$ & $\boldsymbol{a_3}$ & $\boldsymbol{a_4}$ & $\boldsymbol{a_5}$\\
    \hline
    \textbf{OFFLINE} & \cellcolor{green!50}Human & \cellcolor{green!50}Human & \cellcolor{green!50}Human & \cellcolor{green!50}Human & \cellcolor{green!50}Human\\
    \hline
    \textbf{ONLINE} & \cellcolor{green!50}Human & \cellcolor{red!50}Robot & \cellcolor{green!50}Human & \cellcolor{green!50}Human & \cellcolor{red!50}Robot\\
    \hline
    \end{tabular}
    \end{center}
    \vspace{-4 mm}
\end{table}

\begin{figure}[t]
	\centering
	\includegraphics[trim=0.0cm 0.0cm 0.0cm 0.0cm,clip,width=1\linewidth]{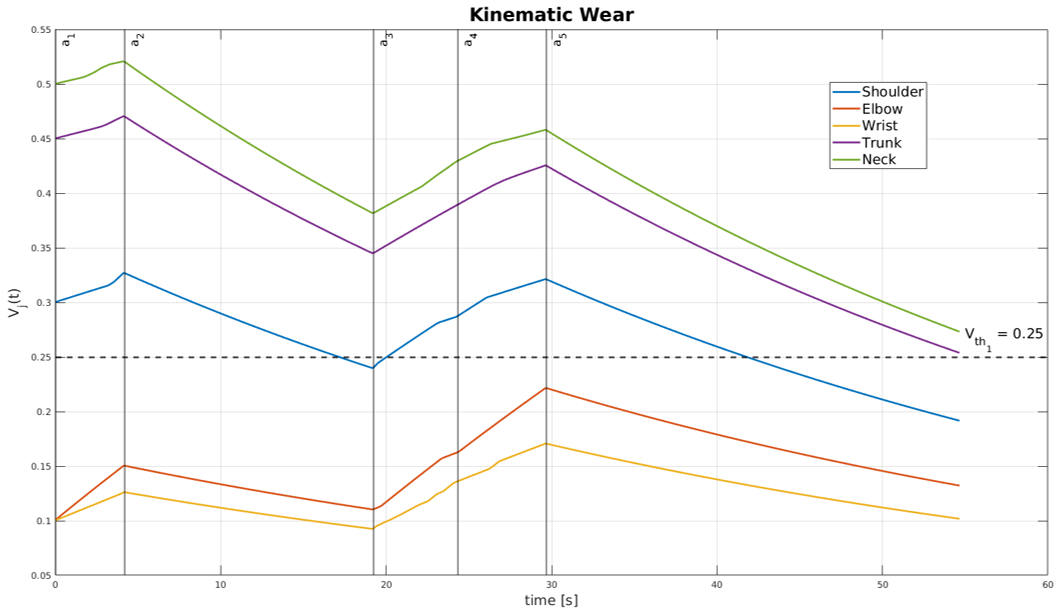}
	\caption{Joint-level \textit{kinematic wear} estimated during the experiment. If the worker does not execute any task, the recovery model is used to lower the wear value that increases during action execution.}
	\label{fig:aog_exp_kin_wear}
    \vspace{-2mm}
\end{figure}
\begin{figure*}[!t]
	\centering
	\includegraphics[trim=0.0cm 0.0cm 3cm 3cm,clip,width=0.19\textwidth]{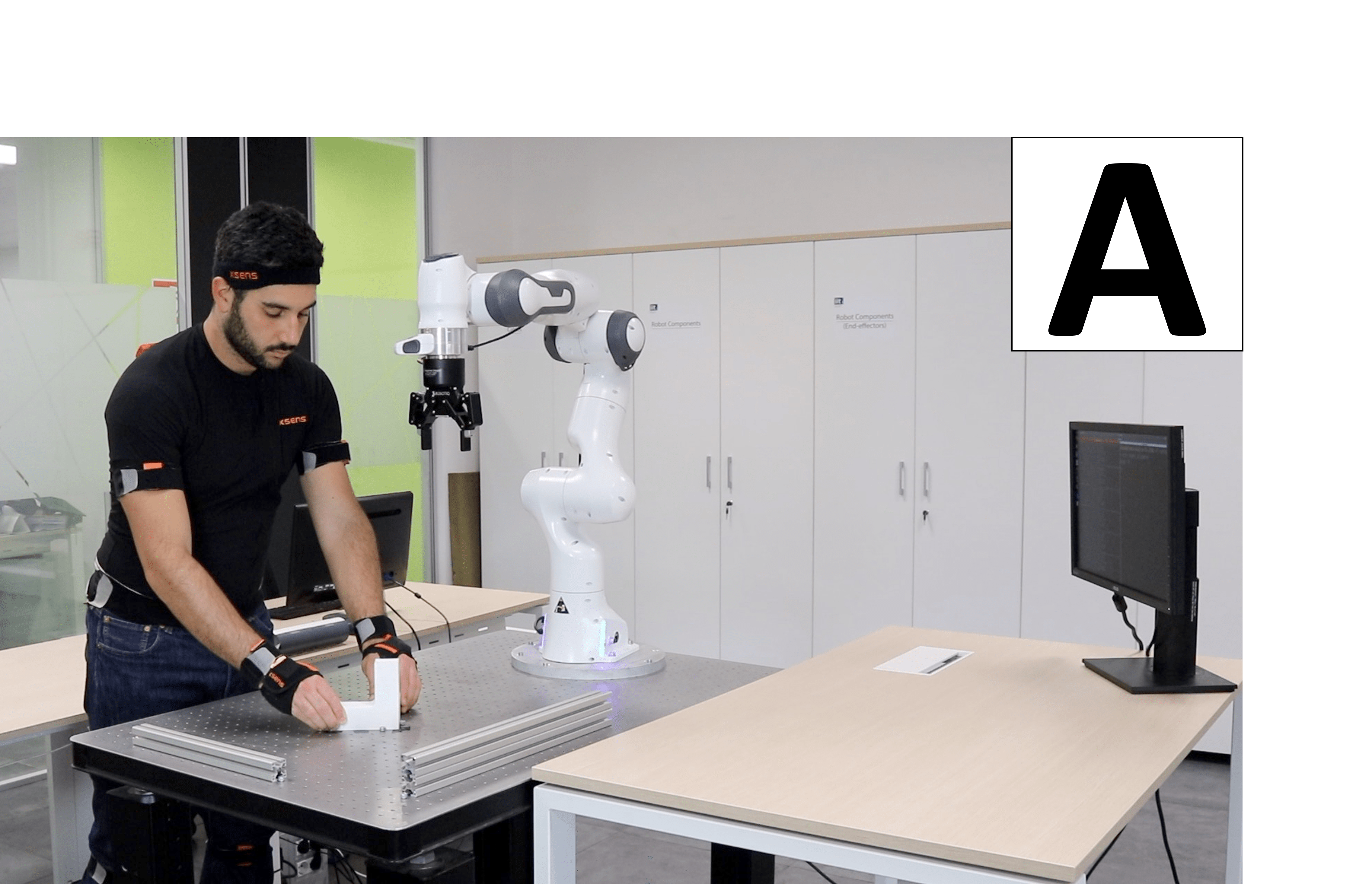}
    \includegraphics[trim=0.0cm 0.0cm 3cm 3cm,clip,width=0.19\textwidth]{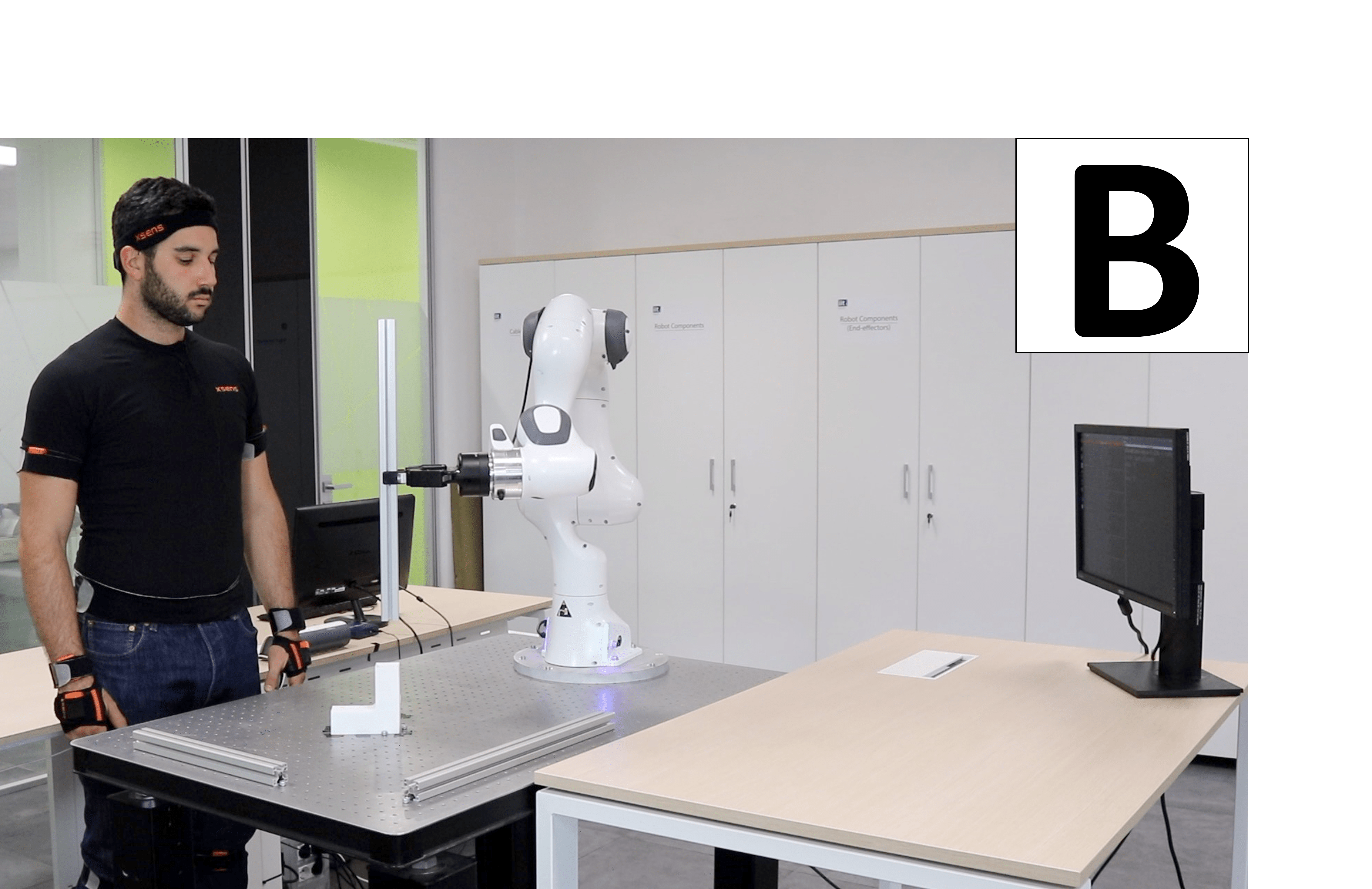}
    \includegraphics[trim=0.0cm 0.0cm 3cm 3cm,clip,width=0.19\textwidth]{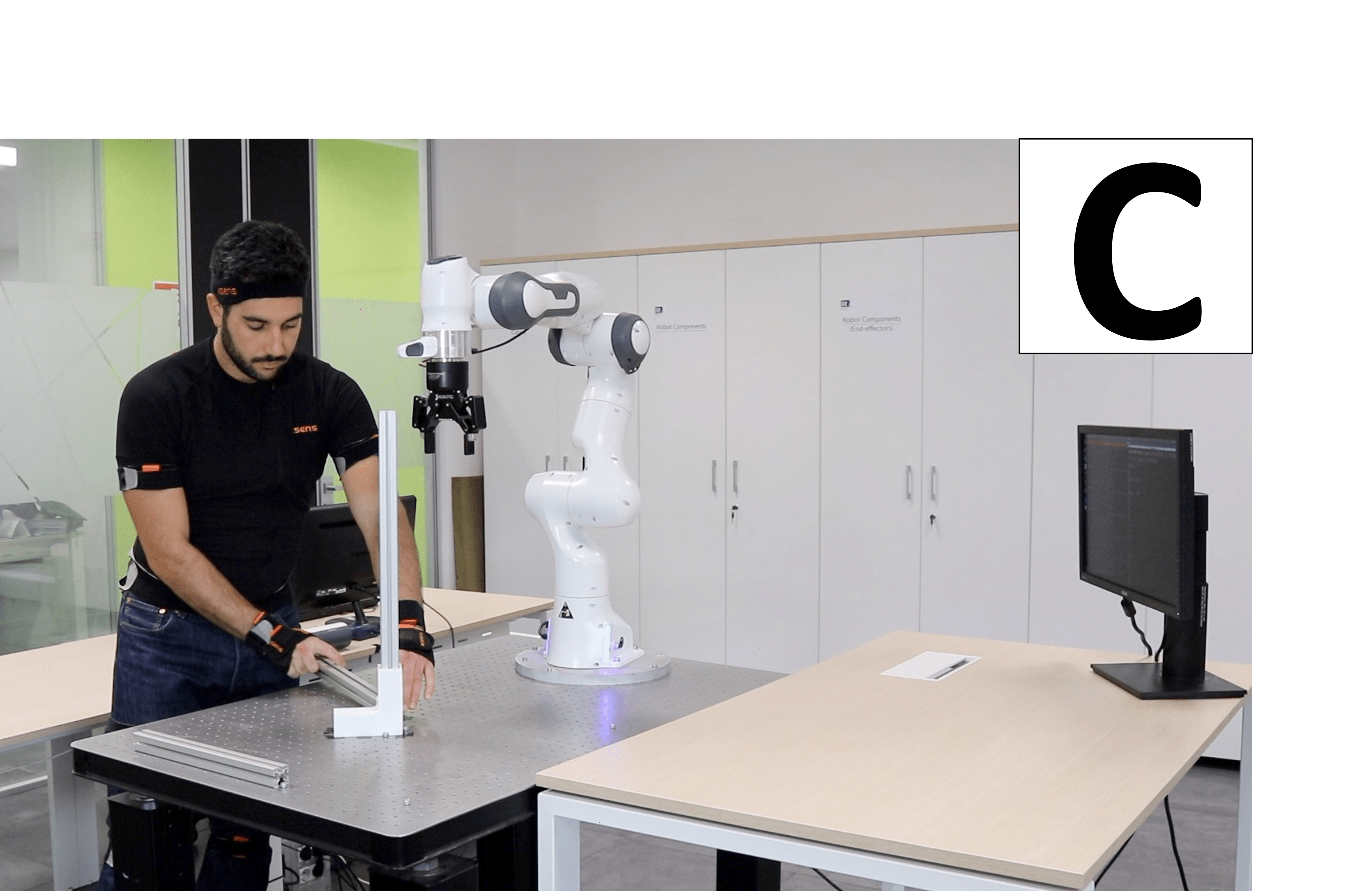}
    \includegraphics[trim=0.0cm 0.0cm 3cm 3cm,clip,width=0.19\textwidth]{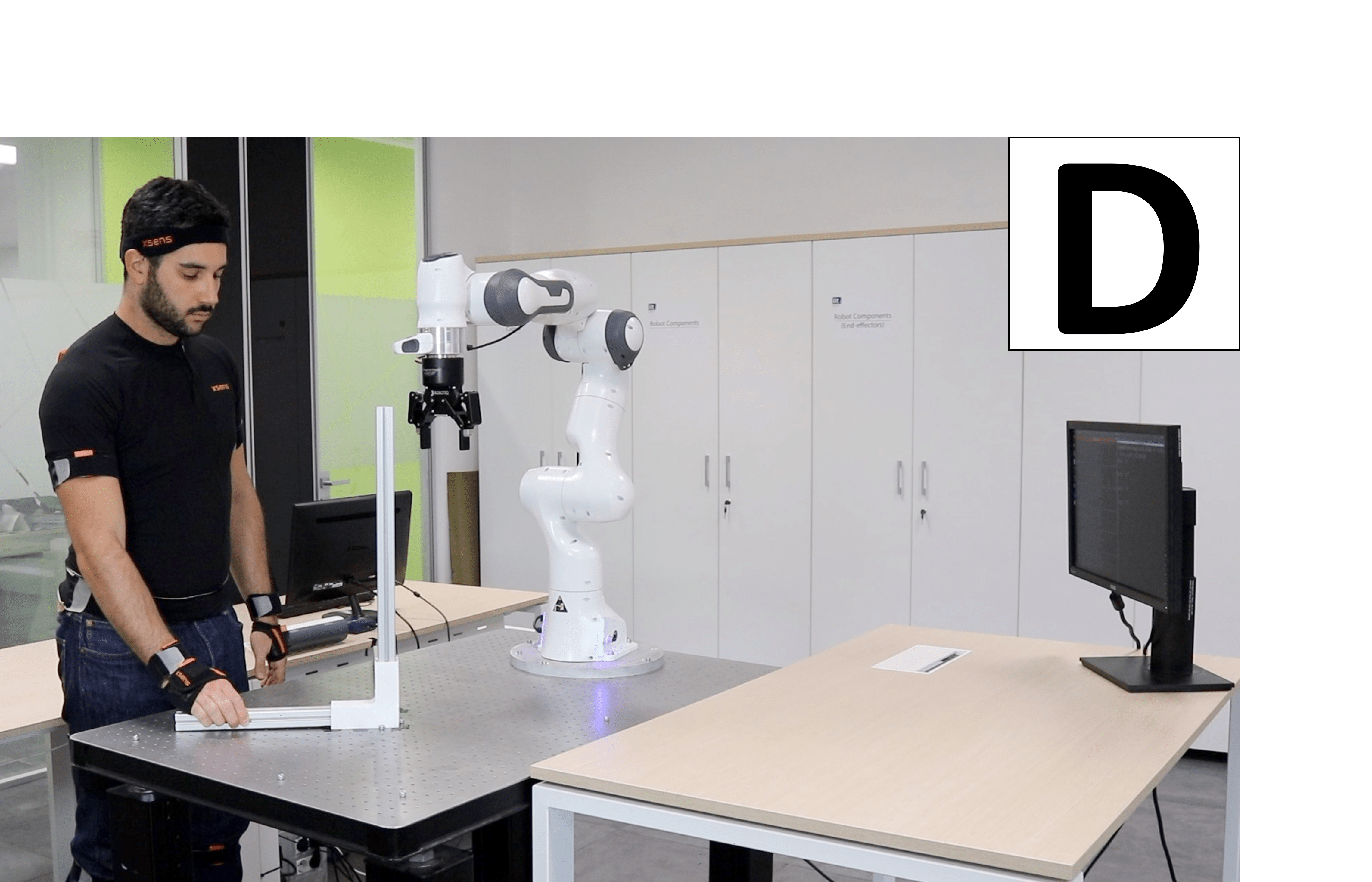}
    \includegraphics[trim=0.0cm 0.0cm 3cm 3cm,clip,width=0.19\textwidth]{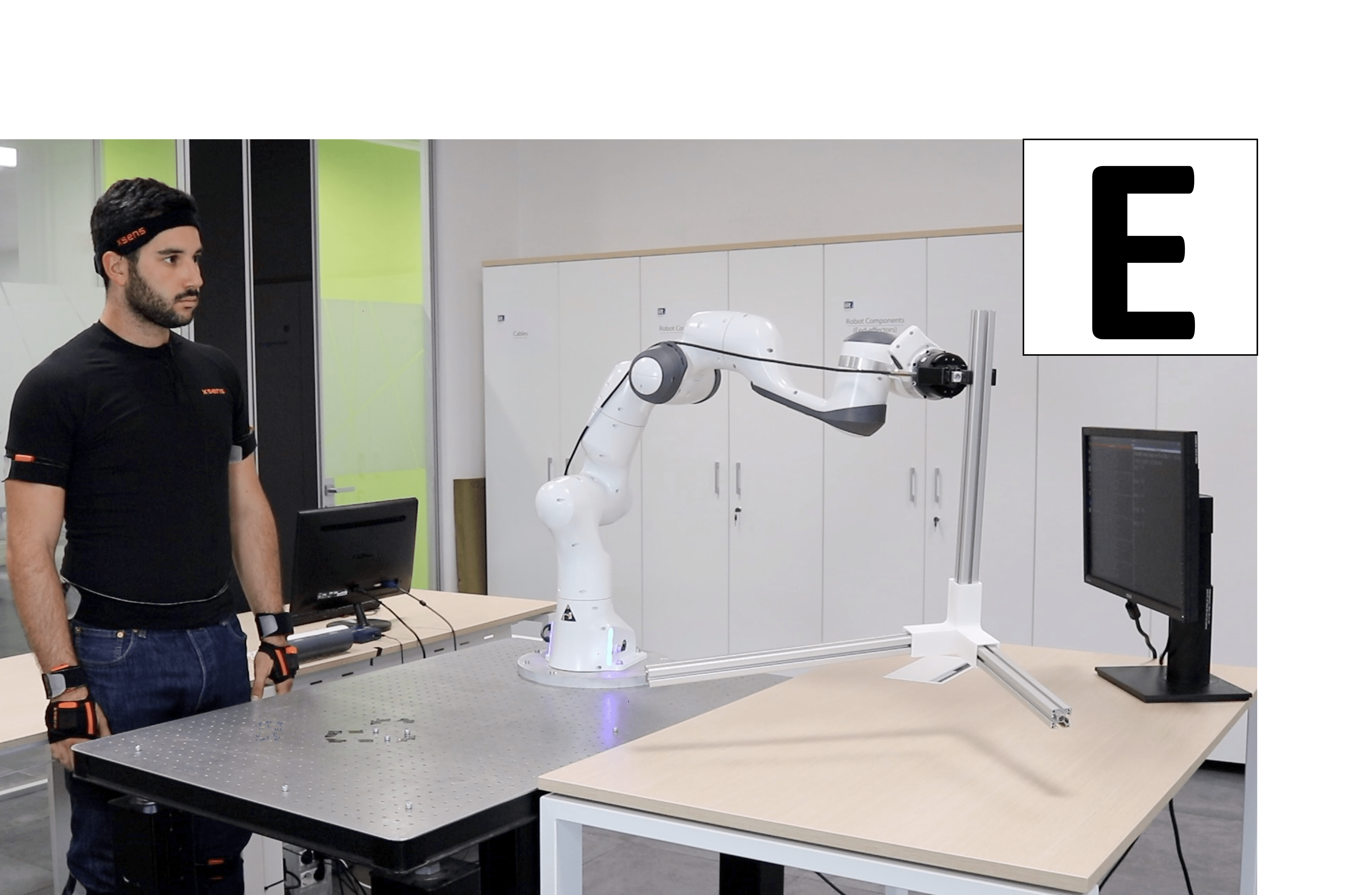}\\
    \includegraphics[trim=0.0cm 0.0cm 0.0cm 0.0cm,clip,width=0.9\linewidth]{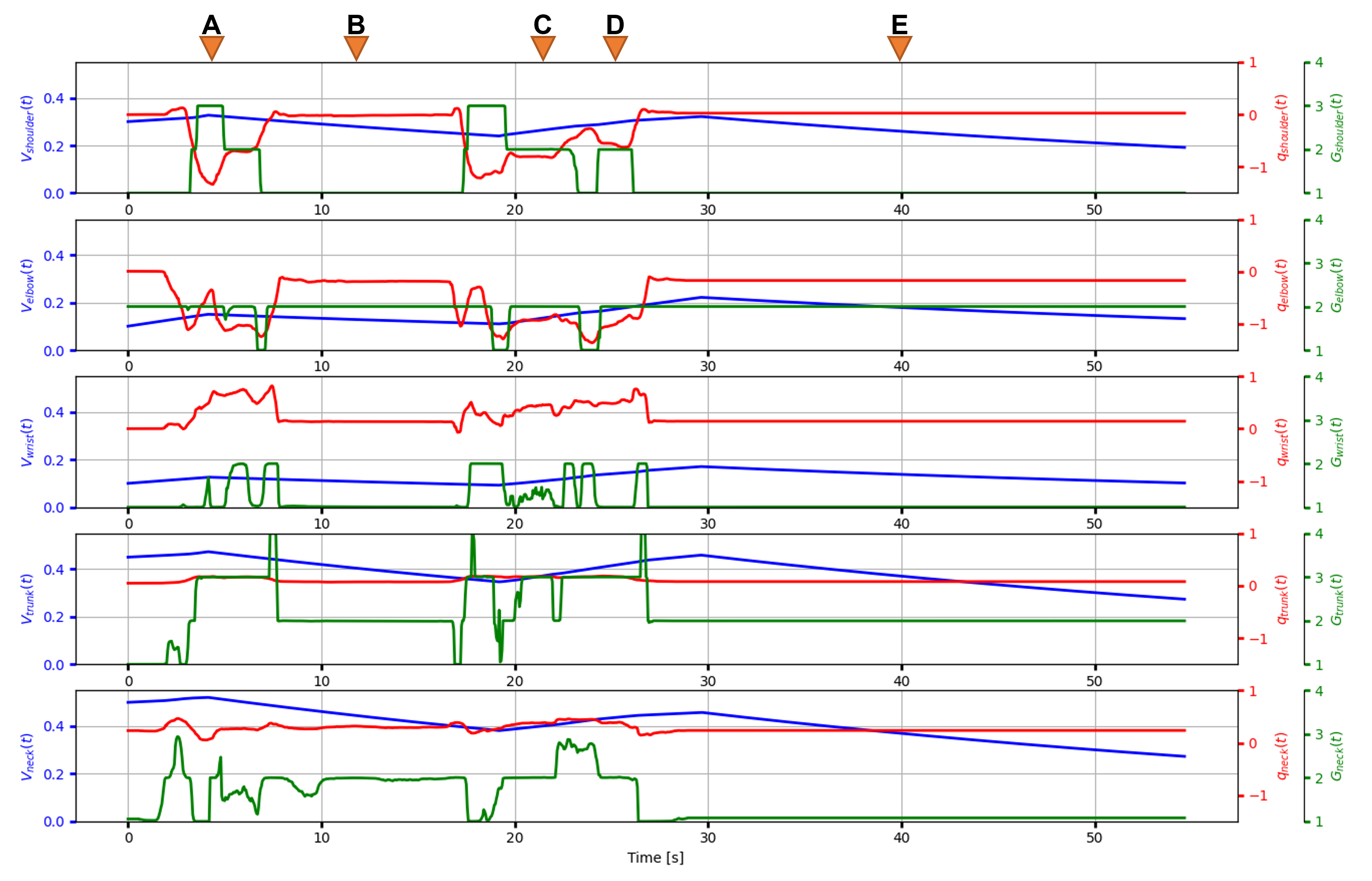}
	\caption{(Top) Snapshots of the experiment. (Bottom) Estimated quantities during the experiment: \textit{kinematic wear} (blue), joint angle in the sagittal plane (red), and the RULA scores (green). A video of the experiments is available in the multimedia extension and in \url{https://youtu.be/OnQLU-4mK_E}.
	}
	\label{fig:aog_exp_allocation_res}
    \vspace{-2mm}
\end{figure*}

The worker is asked to execute the assembly task by following the results of the dynamic allocation method. The \textit{kinematic wear} index for all the joints is non-zero at the beginning of the experiment. In this way, we simulated a scenario where the human worker repetitively executes the same task.
In particular, $V_{shoulder}(0) = 0.3, V_{elbow}(0) = 0.1, V_{wrist}(0) = 0.1, V_{trunk}(0) = 0.45, V_{neck}(0) = 0.5$.
As mentioned beforehand, the actions are performed in the following fixed order: $a_1$, $a_2$, $a_3$, $a_4$, $a_5$. The snapshots of the experiment are present in \autoref{fig:aog_exp_allocation_res}.

The first action is allocated to the human worker: the initial conditions envision three joints (shoulder, trunk, neck) in the medium risk band ($V_{th1} < 0.3, 0.45, 0.5 < V_{th2}$), the other two in the low risk range ($0.1 < V_{th1}$), and the prediction coefficients suggest that the following action would not increase the elbow and wrist \textit{kinematic wear} level enough that $\hat{V}_{elbow, 1} > V_{th1}$ or $\hat{V}_{wrist, 1} > V_{th1}$ (see \autoref{fig:aog_exp_kin_wear}). As a consequence, the cost of action $a_1$ performed by the human is set to $c_H = 32$, against the fixed cost for robot actions $c_R=35$. The algorithm chooses the human worker for $a_1$.
At the end of the first action $V_{shoulder}, V_{trunk}, V_{neck}$ are always in the medium risk zone, while $V_{elbow}$ and $V_{wrist}$ are closer to the lower threshold $V_{th1}$. Due to the values of the kinematic wear along with the prediction coefficients, the algorithm chooses the robot for $a_2$. Allocating another action to the human would entail having more than three joints in the middle range of risk.
While the robot executes action $a_2$, the human can rest and the kinematic wear indexes decrease according to the recovery mode. When the robot finishes $a_2$ the human is again the most suitable agent for performing $a_3$ ($\hat{V}_i < V_{th1} \hspace{0.2cm} \forall{i = 1, \dots, m}$).
At the end of $a_3$, the most stressed joints are the shoulder and the neck. However, the risk level is not high enough to avoid the human execution of the next action, the algorithm assigns $a_4$ to the human worker.
Finally, once $a_4$ has been executed, the predicted values result high enough that the allocation selects the robot for performing $a_5$, and lets the human recover.
By lowering (or increasing) $c_{R}$, it is possible to prioritize the cobot (human) in task allocation procedure. 

The results are compared with a scenario where the kinematic wear is not updated during the task performance (offline assignment). In such case, all the allocations are made according to the risk value estimated for the single task during the calibration process, that does not consider the history of joint usage of the agent. According to such values, the human worker is considered suitable for all the actions (see \autoref{table:allocation_res}).
For these reasons, the offline approach imposes the worker to operate in potentially unhealthy conditions, since it is not aware of the changes of the human physical status during the action execution.

\section{CONCLUSION}
The proposed method envisions a dynamic human-robot role allocation based on an online evaluation of the ergonomic risk. The teamed cooperation uses an adapted AND/OR graph, where the actions, potentially executed by all the agents, are modeled with different hyper-arcs. To each hyper-arc, that represents a human action, an ergonomic cost is assigned. Such costs are updated during the whole duration of the task, modeling the history of the human worker's joints wear. An online AO* search is designed to find the optimal path that consists of the next action and the allocated agent.
While it estimates the ergonomic risk, the method only decides future tasks, but it has the potential also to inform the worker of the risk he/she is exposed to while executing the task. 
To introduce more variability on the demonstrated task, the calibration protocol might involve repetitions of the same task in different tiredness conditions. In this way, it might be possible to obtain a more robust wear prediction and make $\alpha$ more descriptive.
Moreover, it would be significant to execute the same procedure with different subjects, to verify the subject-specificity of the method and investigate algorithms to compute the subject-specific parameters, such as the wear thresholds $V_{th}$.

\balance
\bibliographystyle{IEEEtran}
\bibliography{biblio}

\end{document}